%% file: acl_latex.tex
\pdfoutput=1

\documentclass[11pt]{article}

\usepackage[preprint]{acl}

\usepackage{times}
\usepackage{latexsym}
\usepackage{microtype}
\usepackage{graphicx}
\usepackage{caption}
\usepackage{subcaption}
\usepackage{booktabs} %
\usepackage{tabularx}
\usepackage{amsmath}
\usepackage{xcolor}
\definecolor{TODO}{rgb}{1,0,0} %
\usepackage{amssymb}
\usepackage{mathtools}
\usepackage{amsthm}
\usepackage{lipsum}  
\usepackage{listings}
\usepackage{CJKutf8}
\usepackage{xspace}
\usepackage{multirow}
\usepackage[framemethod=TikZ]{mdframed}
\usepackage{tabu}
\usepackage{longtable}
\usepackage{alphalph}
\usepackage{bm}
\usepackage{tcolorbox}
\usepackage[T1]{fontenc}

\usepackage[utf8]{inputenc}

\usepackage{microtype}

\usepackage{inconsolata}

\usepackage{graphicx}

\renewcommand{\paragraph}[1]{\vspace{0.25em}\noindent\textbf{#1}}

\makeatletter
\def\@secpenalty{-300}
\makeatother
\makeatletter
\renewcommand{\@afterheading}{%
  \@nobreakfalse
  \everypar{%
    \if@nobreak
      \@nobreakfalse
      \clubpenalty 50
    \else
      \clubpenalty \@clubpenalty
      \everypar{}%
    \fi}}
\makeatother

\newcolumntype{C}{>{\centering\arraybackslash}X}

\definecolor{promptbg}{RGB}{243,244,246}   %
\definecolor{promptborder}{RGB}{56,189,248} %

\tcbset{
  promptbox/.style={
    colback=promptbg,
    colframe=promptborder,
    boxrule=0.8pt,
    arc=5pt,
    left=4pt,
    right=4pt,
    top=4pt,
    bottom=4pt,
    fonttitle=\bfseries,
    before skip=10pt, after skip=10pt,
    boxsep=4pt,
    width=\linewidth,
    fontupper=\ttfamily\small,
  }
}

\title{CoVA-SFT: A Large-Scale Dataset for Chain of Visual Abstractions}

\author{\textbf{Tsung-Han Wu}\footnotemark[1]~\quad\textbf{Heekyung Lee}\footnotemark[1]~\quad\textbf{Anya Ji}~\quad\textbf{Haoming Chen}\\
\textbf{Trevor Darrell}\footnotemark[2]~\quad\textbf{Joseph E. Gonzalez}\footnotemark[2]~\quad
\textbf{David M. Chan}\footnotemark[2] \\[8pt]
University of California, Berkeley\\[6pt]
\raisebox{-0.2em}{\includegraphics[height=0.9em]{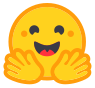}}\
\href{https://huggingface.co/datasets/tsunghanwu/cova}{\small{{\textbf{Dataset: https://huggingface.co/datasets/tsunghanwu/cova}}}}
}

\begin{document}
\maketitle

\footnotetext[1]{*Equal contribution.} \footnotetext[2]{$^\dagger$Equal advising.}

\input{sections/0_abstract}

\input{sections/1_intro_background}
\input{sections/3_methods}
\input{sections/4_experiments}
\input{sections/5_conclusion}

\bibliography{custom}

\clearpage
\appendix

\renewcommand{\theequation}{\thesection.\arabic{equation}}
\renewcommand{\thefigure}{\thesection.\arabic{figure}}
\renewcommand{\thetable}{\thesection.\arabic{table}}

\makeatletter
\@addtoreset{equation}{section}
\@addtoreset{figure}{section}
\@addtoreset{table}{section}
\makeatother

\section*{Appendix}
\label{sec:appendix}
\input{sections/6_appendix}

\end{document}

%% file: sections/0_abstract.tex
\begin{abstract}
Chain-of-thought (CoT) reasoning has dramatically improved large language models (LLMs) by allowing them to decompose problems into intermediate steps. While CoT is widely effective for linguistic tasks, text-only CoT forces models to serialize visual problems into awkward prose. Although architectural solutions exist to process visual inputs, the community lacks a massive, multi-step, self-corrected dataset to teach models \textit{how} to build and maintain internal visual workspaces when solving purely textual reasoning problems. To address this limitation, we introduce \textbf{$\text{CoVA-SFT}$}, a highly structured corpus of 51.9K samples containing over 222K multimodal reasoning steps across 5 distinct layout families and 17 complex tasks, and \textbf{$\text{CoVA-Bench}$}, a companion benchmark of 1,700 held-out test samples spanning the same tasks for reproducible evaluation. By providing explicit rationale formulations, agentic renderings, and verification loops, $\text{CoVA-SFT}$ teaches multimodal language models to interleave text and visual abstractions. We validate the dataset by demonstrating that models fine-tuned on $\text{CoVA-SFT}$ outperform all interleaved CoT baselines by more than $2\times$ on average on $\text{CoVA-Bench}$, though they still fall short of strong text-only CoT baselines, highlighting open challenges for future work.

\end{abstract}

%% file: sections/1_intro_background.tex
\section{Introduction \& Background}

Long chain-of-thought (CoT) reasoning has produced large gains in mathematical problem solving, coding, and other complex tasks \cite{jaech2024openai,guo2025deepseek}, becoming a foundational interface that effectively allows large language models (LLMs) to decompose problems and write intermediate steps before producing an answer. While this interface is effective when the relevant intermediate state is naturally linguistic (such as a mathematical proof), it fails to easily generalize to reasoning problems grounded in non-textual, structured states where serializing into text is inefficient. For example, for humans, reasoning over a picture of an actual chessboard is much easier than interpreting a long algebraic notation of a chess game. Yet, in these cases, text-only chain-of-thought forces models to serialize visual states into an unnatural prosaic representation.

\begin{figure}[t]
    \centering

    \includegraphics[
        width=\linewidth]{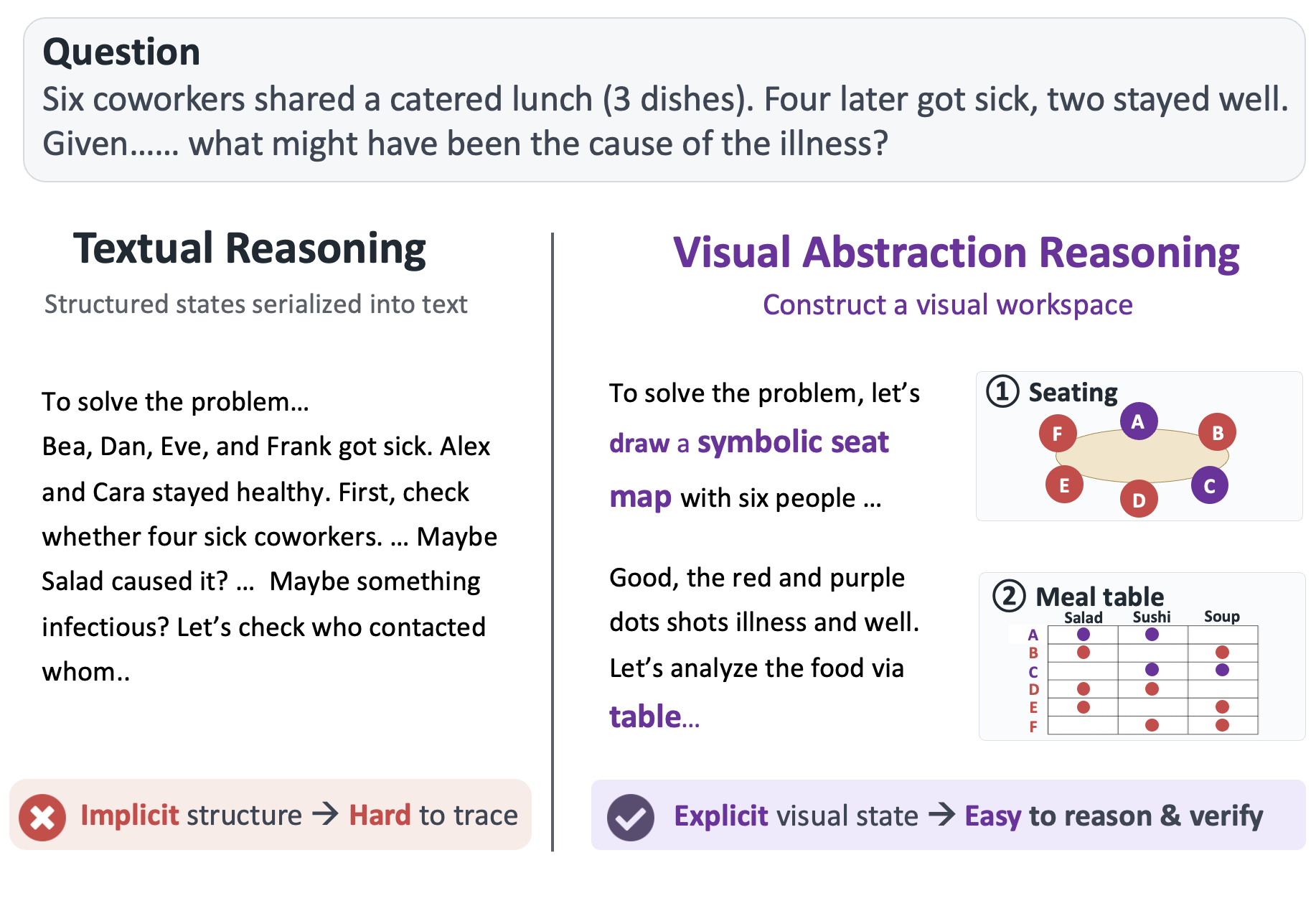}
    
    \caption{\textbf{Chain of Visual Abstractions.} When solving text-based reasoning problems, standard textual reasoning (left) forces models to serialize naturally visual problems into prose. In contrast, models trained on our \textbf{$\text{CoVA-SFT}$} dataset (right) learn to construct and maintain (latent) visual workspaces during the reasoning process.}
\end{figure}

\begin{table*}[t]
\centering
\small
\setlength{\tabcolsep}{4.5pt}
\renewcommand{\arraystretch}{1.15}
\caption{Comparison with representative interleaved reasoning datasets. Prior work grounds reasoning in existing images; $\text{CoVA-SFT}$ constructs visual workspaces from scratch for text-only inputs.}
\label{tab:dataset_comparison}
\resizebox{\linewidth}{!}{
\begin{tabular}{llll}
\toprule
\textbf{Dataset}
& \textbf{Task}
& \textbf{Input}
& \textbf{Visual Role} \\
\midrule

Zebra-CoT
& Vision-language reasoning
& Multimodal
& Grounded visual observations from input \\

Math-VR
& Math reasoning
& Text \& multimodal
& Math-specific plots and diagrams \\

\textbf{CoVA-SFT (Ours)}
& \textbf{Text-only reasoning}
& \textbf{Text-only}
& \textbf{Visual workspaces constructed during reasoning} \\
\bottomrule
\end{tabular}
}
\end{table*}

\begin{figure*}
    \centering

    \includegraphics[
        width=\textwidth,
    ]{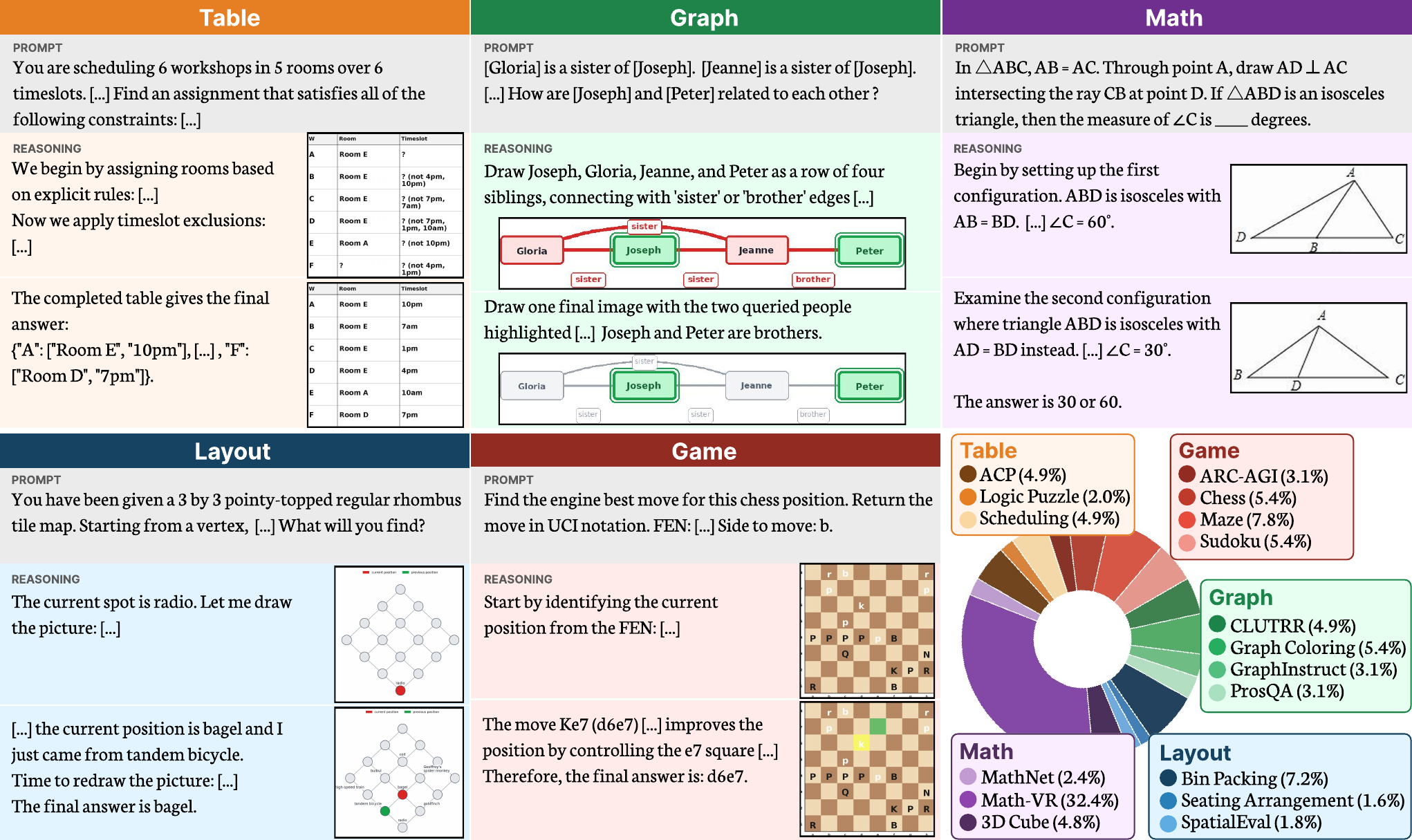}

    \caption{\textbf{Overview of CoVA-SFT.} We introduce CoVA-SFT, a 51.9K-sample dataset containing over 222K images from 17 data sources. CoVA-SFT spans five visual abstraction types, reflecting how humans often externalize reasoning such as tables for constraint problems, family trees for kinship reasoning, graphs for relational reasoning, symbolic diagrams for spatial reasoning, and geometry plots for math reasoning. Data curation pipeline is in \autoref{fig:pipeline} and more dataset examples in 17 tasks are in \autoref{app:data-examples}.}
    \label{fig:dataset-examples}
\end{figure*}

Indeed, recent work has increasingly challenged the assumption that intermediate reasoning must only be expressed in language, extending it beyond text with sketches, generated images, code-rendered plots, and interleaved vision-language traces \cite{hu2024visual,li2025imagine,su2025openthinkimg,deng2025openvlthinker,li2025latent,wang2025autoregressive,bigverdi2025perception}. For example, in visual question answering, latent-token approaches such as Mirage and CoVT replace parts of the textual reasoning trace with continuous visual representations aligned to features coming from input images in the problem \cite{qin2025chain,yang2025machine}. In natural language domains, models can render explicit drawings or executable plotting traces from text queries to use as external scratchpads for spatial or mathematical reasoning \cite{menon2024whiteboard,duan2025codeplot}. Other efforts, like Zebra-CoT, have contributed interleaved vision-language reasoning data \cite{li2025zebracot}.

Despite these advances, the community lacks a massive, multi-step, self-corrected dataset to teach models \textit{how} to build internal visual workspaces. Existing resources are often limited in scale, domain diversity, or the depth of sequential reasoning required. Consequently, researchers studying visual intermediate representations lack a comprehensive benchmark dataset to train and validate models on purely text-based reasoning tasks that require long-horizon structural tracking.

To address this gap, we introduce \textbf{$\text{CoVA-SFT}$}, a highly structured corpus designed to teach multimodal language models to interleave text and visual abstractions. The dataset comprises 51,904 samples containing 222,046 images across 5 distinct layout families (Game, Graph, Layout, Math, and Table) sourced from 17 complex tasks. We construct this dataset using an agentic generation pipeline that explicitly forces models to articulate the rationale for a visual abstraction, agentically render it, and verify its structural consistency in a self-reflective loop. 

To evaluate $\text{CoVA-SFT}$, we also release \textbf{$\text{CoVA-Bench}$}, a held-out benchmark of 1,700 test samples spanning all 17 tasks from the same five layout families, designed to measure logical interleaved-reasoning skills. We fine-tune a base multimodal language model on $\text{CoVA-SFT}$ and evaluate on $\text{CoVA-Bench}$, showing that while strong text-only models like Qwen3-Think struggle significantly on complex domains such as Graph reasoning (44.5\%) and Game simulation (19.3\%), models trained on our dataset can effectively learn to leverage visual abstractions. To sum up, our contributions are:
\begin{itemize}
    \item We release \textbf{$\text{CoVA-SFT}$}, a dataset of 51,904 trajectories with 222,046 visual abstraction steps across 17 tasks, constructed via a self-corrective pipeline.
    \item We provide a dataset validation baseline, confirming that $\text{CoVA-SFT}$ is learnable, challenging, and successfully teaches models to utilize latent visual workspaces without compromising general text reasoning performance.
    \item We release \textbf{$\text{CoVA-Bench}$}, a benchmark consisting of 1,700 test samples, including all 17 tasks of CoVA-SFT that require logical interleaved-reasoning skills.
\end{itemize}

%% file: sections/3_methods.tex
\section{Interleaved Reasoning Dataset: CoVA-SFT}
\label{sec:methods}

To evaluate the potential of intermediate visual reasoning, we introduce \textbf{$\text{CoVA-SFT}$}, a multi-source interleaved reasoning corpus of 51,904 examples. The dataset contains 222,046 multimodal reasoning steps across 17 diverse tasks, providing a set of trajectories that teach models how to build internal workspaces when solving complex textual reasoning problems.

\begin{figure}
    \centering

    \includegraphics[
        width=0.9\linewidth,
    ]{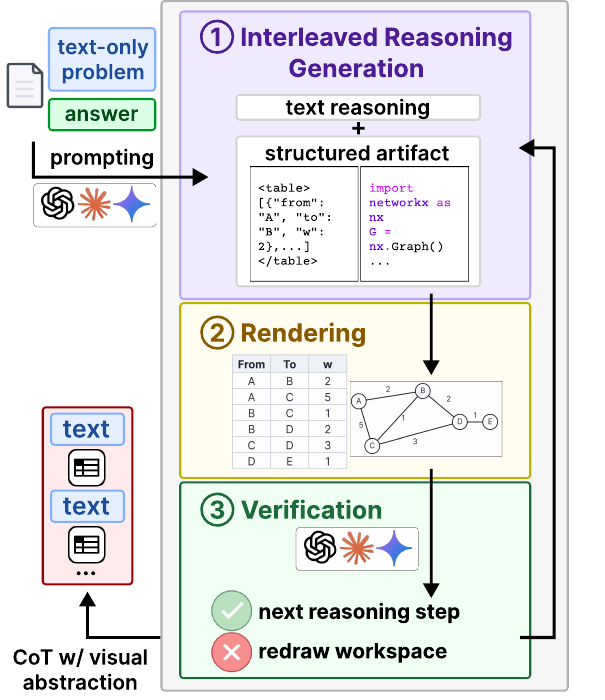}

    \caption{\textbf{Data synthesis pipeline.} A capable vision-language model generates text reasoning traces and structured intermediate artifacts, renders them into visual workspaces, and verifies the resulting outputs through a reflective feedback loop before continuing downstream reasoning.}
    \label{fig:pipeline}
\end{figure}

\input{table/main}

\paragraph{Dataset Construction.}
We construct $\text{CoVA-SFT}$ using Claude 4.5 Sonnet as an agentic generator operating in three stages (\autoref{fig:pipeline}):
\begin{enumerate}
    \item \textbf{Rationale \& QA Formulation}: The model articulates \textit{why} a visual workspace is useful for the given task, specifies what should be rendered, and generates the corresponding question--answer pair.
    \item \textbf{Agentic Rendering}: The interleaved reasoning trace is generated step by step, with Matplotlib tool calls to programmatically render each visual workspace; the resulting image is fed back to the model as context before continuing.
    \item \textbf{Verification \& Self-Correction}: The model checks the rendered image for structural consistency against the problem statement and re-enters the rendering loop to redraw when errors are detected.
\end{enumerate}
The dataset spans five abstraction families (Table, Graph, Layout, Game, and Math) covering 17 tasks in total. In addition to the training set, we also curate the \textbf{$\text{CoVA-Bench}$}, a held-out test benchmark drawn from the same 17 tasks and five abstraction families as $\text{CoVA-SFT}$. It contains 1,700 samples (100 per task) that were withheld from training and verified to require multi-step logical reasoning and maintenance of visual state. 

%% file: table/main.tex
\begin{table*}
\centering
\small
\setlength{\tabcolsep}{5pt}
\renewcommand{\arraystretch}{1.15}
\caption{
Dataset Validation Baseline. We compare text-only baselines and external-tool visual reasoning methods against the CoVA-SFT baseline. Text-only serialization heavily degrades performance on Graph and Game tasks.
}
\label{tab:eval_overview}
\begin{tabularx}{\linewidth}{llXcccccc}

\toprule

\textbf{CoT Format} & \textbf{Instantiation} & \textbf{Method}

& \textbf{Table} & \textbf{Layout} & \textbf{Graph} & \textbf{Game}

& \textbf{Math} & \textbf{Avg} \\

\midrule

\multirow{3}{*}{\textcolor{gray}{Text-Only}}

& \textcolor{gray}{LLM}

& \textcolor{gray}{Qwen3-Think} & \textcolor{gray}{67.3} & \textcolor{gray}{69.3} & \textcolor{gray}{44.5} & \textcolor{gray}{19.3} & \textcolor{gray}{25.0} & \textcolor{gray}{45.1} \\

\cmidrule(lr){2-9}

& \multirow{2}{*}{\textcolor{gray}{VLM}}

& \textcolor{gray}{Qwen3-VL-Instruct} & \textcolor{gray}{61.0} & \textcolor{gray}{71.0} & \textcolor{gray}{57.0} & \textcolor{gray}{15.0} & \textcolor{gray}{26.0} & \textcolor{gray}{46.0} \\

&

& \textcolor{gray}{Qwen3-VL-Thinking} & \textcolor{gray}{80.3} & \textcolor{gray}{76.3} & \textcolor{gray}{46.5} & \textcolor{gray}{25.5} & \textcolor{gray}{27.3} & \textcolor{gray}{51.2} \\

\midrule

\multirow{5}{*}{Interleaved}

& \multirow{4}{*}{Agentic / UMM}

& CodePlot-CoT~\cite{duan2025codeplot} & 15.3 & 17.3 & 20.2 & \ \ 2.0 & \ \ 9.7 & 12.9 \\

&

& TwGI~\cite{chern2025thinking} & \ \ 2.0 & \ \ 0.0 & \ \ 1.0 & \ \ 0.0 & \ \ 1.3 & \ \ 0.9 \\

&

& MathCanvas~\cite{shi2025mathcanvas} & 14.0 & 22.0 & 35.3 & \ \ 0.5 & \textbf{12.3} & 16.8 \\

&

& Zebra-CoT~\cite{li2025zebracot} & \ \ 9.3 & 13.8 & 26.5 & \ \ 0.5 & \ \ 5.7 & 11.2 \\

\cmidrule(lr){2-9}

& Latent Tokens

& \textbf{CoVA-SFT Baseline (Ours)}

& \textbf{47.2} & \textbf{53.4} & \textbf{62.0} & \textbf{20.0} & \ \ 8.5 & \textbf{38.2} \\

\bottomrule

\end{tabularx}
\end{table*}

%% file: sections/4_experiments.tex
\section{Experiments}
\label{sec:experiments}

To validate that $\text{CoVA-SFT}$ is learnable and that interleaved visual reasoning provides a meaningful training signal, we fine-tune a multimodal language model on $\text{CoVA-SFT}$ and evaluate on \textbf{$\text{CoVA-Bench}$}, our held-out benchmark of 1,700 test samples across all 17 tasks. We compare against text-only and interleaved CoT baselines under zero-shot conditions, where each model receives only the textual problem statement.

\paragraph{Model Design.}
We fine-tune Qwen3-VL-8B-Thinking \cite{bai2025qwen3} on $\text{CoVA-SFT}$, adapting it to our interleaved reasoning format where the model must produce not only the final textual answer but also the intermediate latent visual trajectory that grounds the reasoning process. Each interleaved image is encoded using a fixed budget of 128 visual tokens. We optimize with a dual-objective loss: for text, the standard autoregressive cross-entropy $\mathcal{L}_{\text{text}}$; for visual tokens, a cosine similarity objective that aligns each normalized predicted latent visual token $e_i$ against its normalized ground-truth target embedding $\hat{e}_i$,
\begin{equation}
    \mathcal{L}_{\text{visual}}
    =
    \frac{1}{N}
    \sum_{i=1}^{N}
    l_{\text{cos}}(e_i, \hat{e}_i),
\end{equation}
yielding the total objective $\mathcal{L}_{\text{SFT}} = \mathcal{L}_{\text{visual}} + \gamma\,\mathcal{L}_{\text{text}}$. This formulation, following \citet{zhang2025openmmreasoner}, explicitly supervises both language generation and visual-state evolution, encouraging the model to maintain grounded multimodal representations throughout long reasoning horizons.

\paragraph{Baselines.}
We evaluate against two categories, selecting open-source models in the 7--8B parameter range to ensure fair comparison under matched capacity. \textit{Text-only CoT} baselines include pure LLMs, Qwen3-Think \cite{yang2025qwen3}, as well as VLMs operated in text-only mode: Qwen3-VL-Instruct and Qwen3-VL-Thinking \cite{bai2025qwen3}. \textit{Interleaved CoT} baselines include methods that produce explicit visual workspaces via external tool calls: CodePlot-CoT \cite{duan2025codeplot}, MathCanvas \cite{shi2025mathcanvas}, and Zebra-CoT \cite{li2025zebracot}, and Thinking with Generated Images \cite{chern2025thinking}, all of which programmatically render or generate visual artifacts and feed them back into the reasoning chain.

\paragraph{Results and Analysis.}
\autoref{tab:eval_overview} reports performance across our in-domain held-out test sets. CoVA-SFT achieves the highest average score among interleaved CoT baselines by a substantial margin (38.2\% vs.\ 16.8\% for the next-best baseline, MathCanvas), although MathCanvas performs better on Math. CodePlot-CoT, MathCanvas, and Zebra-CoT remain substantially weaker overall, as errors in external rendering can propagate into subsequent reasoning with limited opportunity for recovery. By internalizing the visual workspace into the reasoning process, CoVA-SFT avoids this external execution boundary. TwGI \cite{chern2025thinking} achieves near-zero performance across domains, as its generated intermediate images are often incoherent and provide little useful grounding over long CoT trajectories.

Against text-only baselines, CoVA-SFT trails on Table (47.2\%), Layout (53.4\%), and Math (8.5\%), but surpasses all text-only models on \textbf{Graph reasoning} (62.0\% vs.\ 57.0\%). This advantage is structurally motivated: graphs encode relational information (nodes, edges, reachability) in a form that is naturally visual but laborious to maintain in prose, where a text-only chain-of-thought must re-enumerate adjacency structure at every step. The sharpest gap is on Math (8.5\% vs.\ 27.3\%), which we attribute to a representational mismatch: mathematical reasoning is already well-served by symbolic notation that VLMs handle fluently in text, making latent visual tokens an expensive detour rather than a meaningful aid. This suggests that learning continuous latent visual representations may be a more effective inductive bias than discrete token generation in unified multimodal models.

%% file: sections/5_conclusion.tex
\section{Conclusion}

In this work, we introduce \textbf{$\text{CoVA-SFT}$}, a large-scale interleaved reasoning dataset designed to address the fundamental limitations of text-only Chain-of-Thought on spatial, structural, and relational tasks. By creating 51,904 trajectories with 222,046 visual abstraction steps across 17 distinct tasks, we hope to provide the community with a resource for teaching multimodal models how to build and maintain internal visual workspaces.

\clearpage
\section*{Limitations}
\label{subsec:limitations}

A primary limitation of $\text{CoVA-SFT}$ is its reliance on the capabilities of the upstream vision-language model (e.g., Claude 4.5 Sonnet) acting as the agentic generator. While our verification and self-correction loops catch many structural inconsistencies, subtle hallucinations in the generated text traces or rendering code can propagate into the training data. Because these trajectories are directly distilled into the dataset, these micro-errors can introduce logic flaws into the training distribution, potentially bounding the upper-limit accuracy of downstream fine-tuned models.

Furthermore, the dataset's visual diversity is currently bounded by the programmatic capabilities of the rendering tools; highly complex, open-world spatial simulations or continuous robotic environments are not yet represented in this taxonomy. Currently, the visual abstractions are constrained to static structured formats such as 2D tables, topological graphs, layout grids, and geometric coordinate plots. Consequently, the visual abstractions learned via $\text{CoVA-SFT}$ remain symbolic and schematic rather than continuous or deeply perceptual.

Finally, in addition to data issues, training models to jointly optimize for both textual tokens and continuous or latent visual tokens introduces significant computational overhead compared to standard text-only Supervised Fine-Tuning (SFT). The dual-objective loss function, which forces the model to balance cross-entropy text generation with high-dimensional cosine similarity alignment ($\mathcal{L}_{SFT}=\mathcal{L}_{visual}+\gamma\mathcal{L}_{text}$), requires careful hyperparameter tuning and increased memory bandwidth during training. Furthermore, because highly complex tasks in our dataset scale up to 8 or more sequential visual states, generation during inference demands longer context windows and an increased compute budget to dynamically update and maintain the workspace over long horizons.

\section*{Acknowledgments} 

Sky Computing Lab is supported by gifts from Accenture, AMD, Anyscale, Cisco, Google, IBM, Intel, Intesa Sanpaolo, Lambda, Lightspeed, Mibura, Microsoft, NVIDIA, Samsung SDS, and SAP.

\vfill

%% file: sections/6_appendix.tex
The appendix is organized as follows:
\begin{itemize}
    \item \autoref{app:results} provides additional performance and evaluation details, including per-task results across the five workspace categories.
    \item \autoref{app:data} describes the construction of the CoVA-SFT dataset, including programmatically generated examples and examples collected from existing open-source datasets.
    \item \autoref{app:data-examples} presents representative training data examples across the supported reasoning tasks and visual abstraction types.
    \item \autoref{app:ethics} discusses ethics, risks, intended use, artifact licensing, and dataset documentation.
    \item \autoref{sec:appendix_hyperparams} discusses training hyperparameters for the CoVA baseline.
    \item \autoref{app:ai} discloses the use of AI-based tools during manuscript preparation.
\end{itemize}

\section{Performance and Evaluation Details}
\label{app:results}

We additionally report detailed per-task performance results for a more fine-grained evaluation of model capabilities across diverse reasoning domains. The results can be found in \autoref{tab:workspace_results}. Ground truth answers in our test set span diverse formats, including letters, words, names, structured tables, and symbolic expressions. We report the main numbers in the paper using the LLM as a judge (Gemini-2.5-flash) setting, as it provides a more reliable estimate of practical reasoning performance.

\section{Dataset Construction Details}
\label{app:data}
CoVA-SFT question-answer pairs are either programmatically generated or collected from existing open-source datasets. Below we describe each source.

\vspace{1em}

\noindent\textbf{Seating Arrangement:}
Tasks are generated using a constraint-satisfaction algorithm over two layout types (straight-line and circular) with 18 seating constraint templates. We sample a random number of seats and iteratively add constraints until a unique valid arrangement remains, discarding any constraint that conflicts with the existing set. Multiple-choice questions are then generated by prompting an LLM on the initial and final arrangements.

\vspace{1em}

\noindent\textbf{Logic Puzzles:}
Generated from the open-source Zebra Puzzles repository,\footnote{\url{https://github.com/alexandrainst/zebra_puzzles}} the pipeline samples a ground-truth world of $n$ houses and $m$ attributes, then adds weighted clue constraints until a CSP solver yields a unique solution. Each puzzle is rendered into natural language with a JSON answer key.

\vspace{1em}

\noindent\textbf{3D Cube:}
Connected voxel structures of varying dimensions are procedurally sampled and rendered into 3D visualizations. Tasks cover three question types: total exposed surface area, cubes with a given number of painted faces, and surface-area change after removing a highlighted cube. Ground-truth answers are computed automatically from the voxel geometry.

\vspace{1em}

\noindent\textbf{Others:} All remaining data is sourced from existing open-source datasets. We provide source details in \autoref{tab:dataset_sources}.
\begin{table*}[t]
\centering
\small
\renewcommand{\arraystretch}{1.15}
\setlength{\tabcolsep}{6pt}
\begin{tabular}{lll}
\toprule
\textbf{Source Repository / URL} \\
\midrule
ARC Interleaved
& \url{https://huggingface.co/datasets/multimodal-reasoning-lab/Zebra-CoT} \\
GraphInstruct
& \url{https://huggingface.co/datasets/GraphWiz/GraphInstruct} \\
CLUTRR
& \url{https://huggingface.co/datasets/CLUTRR/v1} \\
SpatialEval
& \url{https://huggingface.co/datasets/yyamada/SpatialEvalLLM} \\
ProsQA
& 
\url{https://github.com/facebookresearch/coconut/tree/main/data}\\
Math-VR
& \url{https://huggingface.co/datasets/gogoduan/Math-VR-train} \\
MathNet
& \url{https://huggingface.co/datasets/ShadenA/MathNet} \\
ACP Bench
& \url{https://huggingface.co/datasets/ibm-research/acp_bench} \\
Bin Packing
& \url{https://huggingface.co/datasets/mideavalwisard/ACCORD} \\
Scheduling CSP
& \url{https://huggingface.co/datasets/strickvl/constraint-sat-1000} \\
GRAM Graph Coloring
& \url{https://huggingface.co/datasets/brozonoyer/gram-graph-coloring} \\
Puzzle-Bench Sudoku
& \url{https://huggingface.co/datasets/zeyuzy/puzzle-bench} \\
2D Maze Dataset
& \url{https://huggingface.co/datasets/achinta3/2d_maze_dataset_with_grid_size} \\
Chess Dataset
& \url{https://huggingface.co/datasets/aac43t34ty34ty34/chess-dataset} \\
\bottomrule
\end{tabular}
\caption{Source datasets used for constructing the interleaved reasoning corpus.}
\label{tab:dataset_sources}
\end{table*}

\begin{table*}[t]
\centering
\small
\renewcommand{\arraystretch}{1.15}
\setlength{\tabcolsep}{8pt}
\begin{tabular}{llcccc}
\toprule
\textbf{Workspace} & \textbf{Method}
& \multicolumn{4}{c}{\textbf{Task type}} \\
\midrule

\multirow{5}{*}{Game}

& 
& ARC & Chess & Maze & Sudoku \\

& Qwen3-Think~\cite{yang2025qwen3}
& 7.0 & 7.0 & 48.0 & 15.0 \\

& Qwen3-VL-Thinking~\cite{bai2025qwen3}
& 23.0 & 9.0  & 50.0  & 20.0  \\

& CoVA-SFT (Ours)
& 3.3 & 0.0 & 77.0 & 0.0 \\

\midrule

\multirow{5}{*}{Graph}

& 
& CLUTRR & GraphColor & GraphInst. & ProsQA \\

& Qwen3-Think~\cite{yang2025qwen3}
& 37.0 & 21.0 & 97.0 & 23.0 \\

& Qwen3-VL-Thinking~\cite{bai2025qwen3}
& 39.0 & 16.0 & 100.0  & 31.0  \\

& CoVA-SFT (Ours)
& 83.6 & 12.5 & 82.0 & 93.8 \\

\midrule

\multirow{5}{*}{Layout}

& 
& BinPack & SeatingArr & SpatialEval & --- \\

& Qwen3-Think~\cite{yang2025qwen3}
& 60.0 & 70.0 & 78.0 & --- \\

& Qwen3-VL-Thinking~\cite{bai2025qwen3}
& 67.0  & 78.0 & 84.0  & --- \\

& CoVA-SFT (Ours)
& 37.5 & 49.2 & 73.4 & --- \\

\midrule

\multirow{5}{*}{Math}

& 
& 3D-Cube & MathVR & MathNet & --- \\

& Qwen3-Think~\cite{yang2025qwen3}
& 18.0 & 25.0 & 32.0 & --- \\

& Qwen3-VL-Thinking~\cite{bai2025qwen3}
& 19.0  & 31.0  & 32.0 & --- \\

& CoVA-SFT (Ours)
& 9.8 & 10.9 & 4.9 & --- \\

\midrule

\multirow{5}{*}{Table}

& 
& ACP & LogicPuzzle & Scheduling & --- \\

& Qwen3-Think~\cite{yang2025qwen3}
& 60.0 & 49.0 & 93.0 & --- \\

& Qwen3-VL-Thinking~\cite{bai2025qwen3}
& 62.0 & 84.0  & 95.0 & --- \\

& CoVA-SFT (Ours)
& 56.2 & 50.8 & 34.5 & --- \\

\bottomrule
\end{tabular}

\caption{Per-category reasoning performance (\%).}
\label{tab:workspace_results}

\end{table*}

\section{Training Data Examples} 
We provide representative interleaved reasoning trajectories from $\text{CoVA-SFT}$ in \autoref{app-fig:dataset-example}, covering all 17 tasks across the five visual abstraction families.

\label{app:data-examples}
\begin{figure*}[p]
    \centering

    \includegraphics[
        width=\textwidth,
        clip,
        trim=0 8cm 0 0
    ]{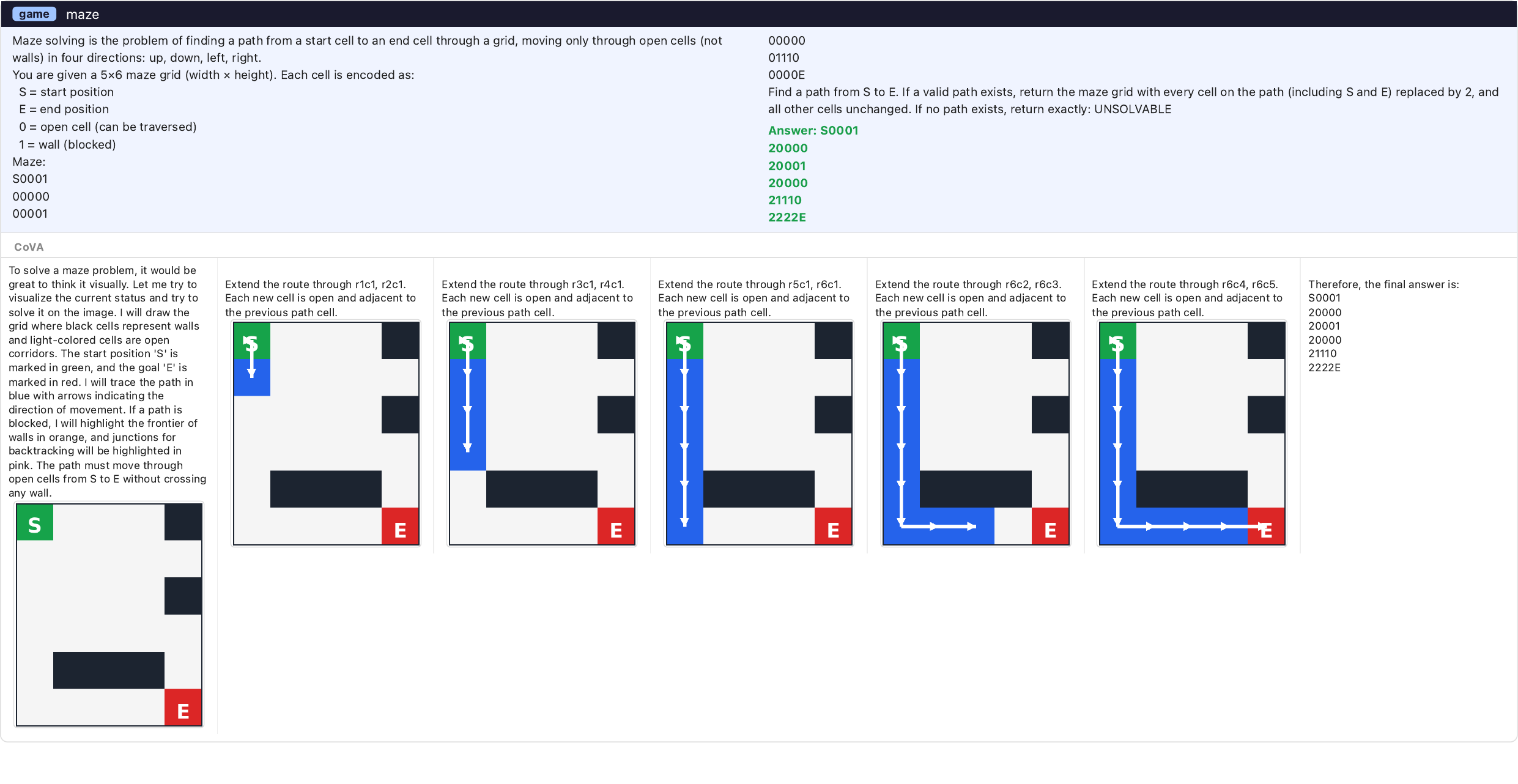}

    \includegraphics[
        width=\textwidth,
        clip,
        trim=0 8.5cm 0 0
    ]{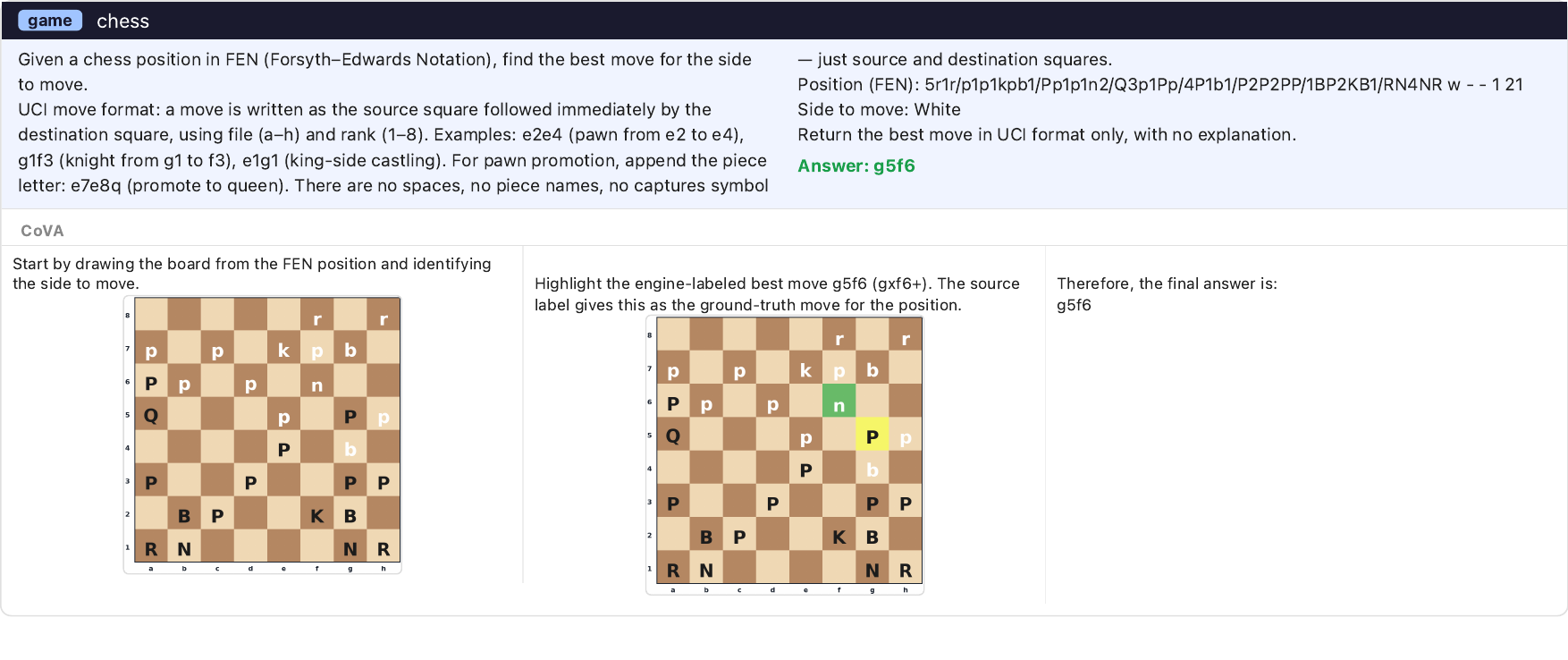}

    \caption{Training data examples.}
\end{figure*}

\begin{figure*}[p]\ContinuedFloat
    \centering

    \includegraphics[
        width=\textwidth,
        clip,
        trim=0 8cm 0 0
    ]{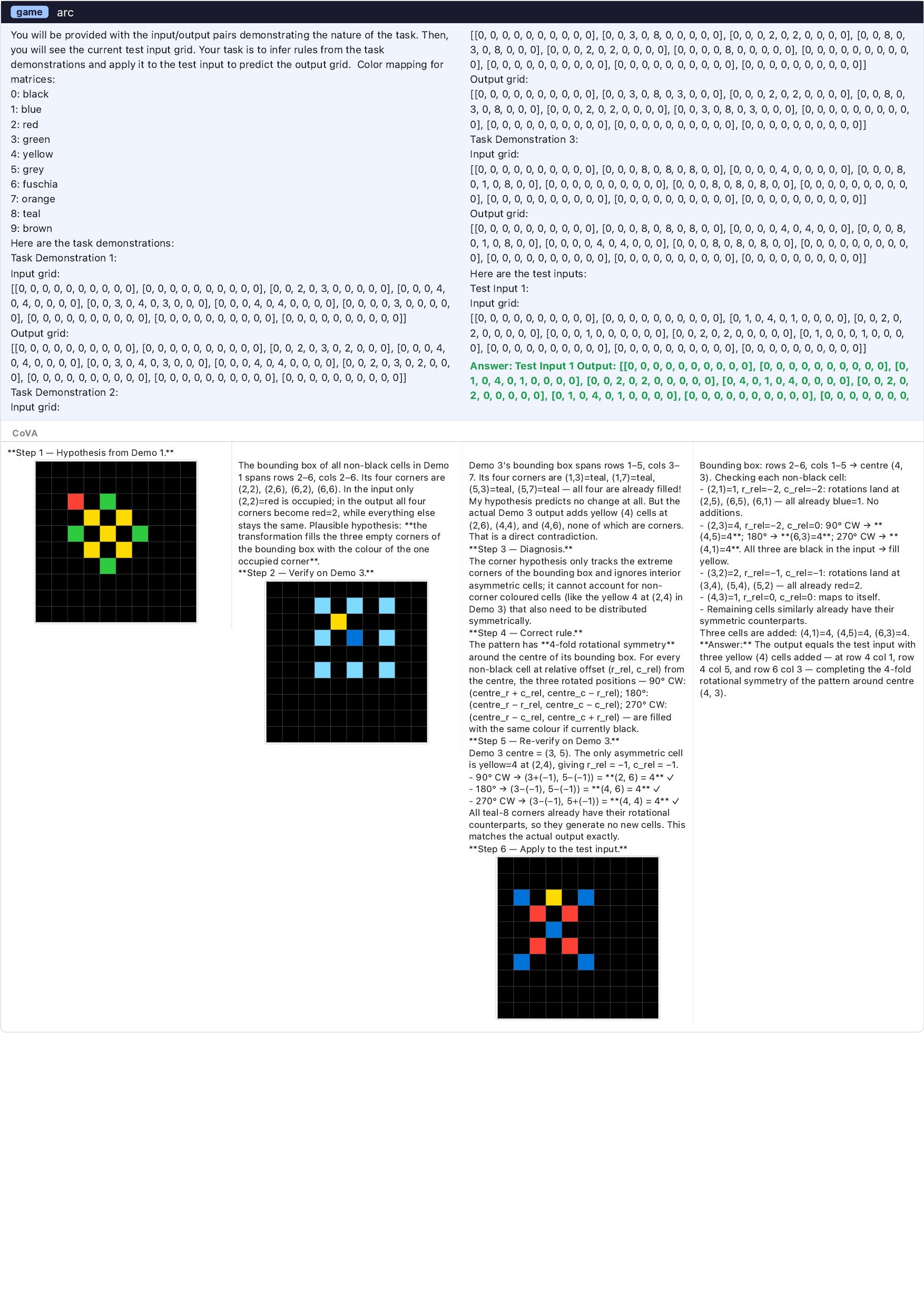}

    \caption{Training data examples (continued).}
    \label{app-fig:dataset-example}
\end{figure*}

\begin{figure*}[p]\ContinuedFloat
    \centering

    \includegraphics[
        width=\textwidth,
        clip,
        trim=0 10cm 0 0
    ]{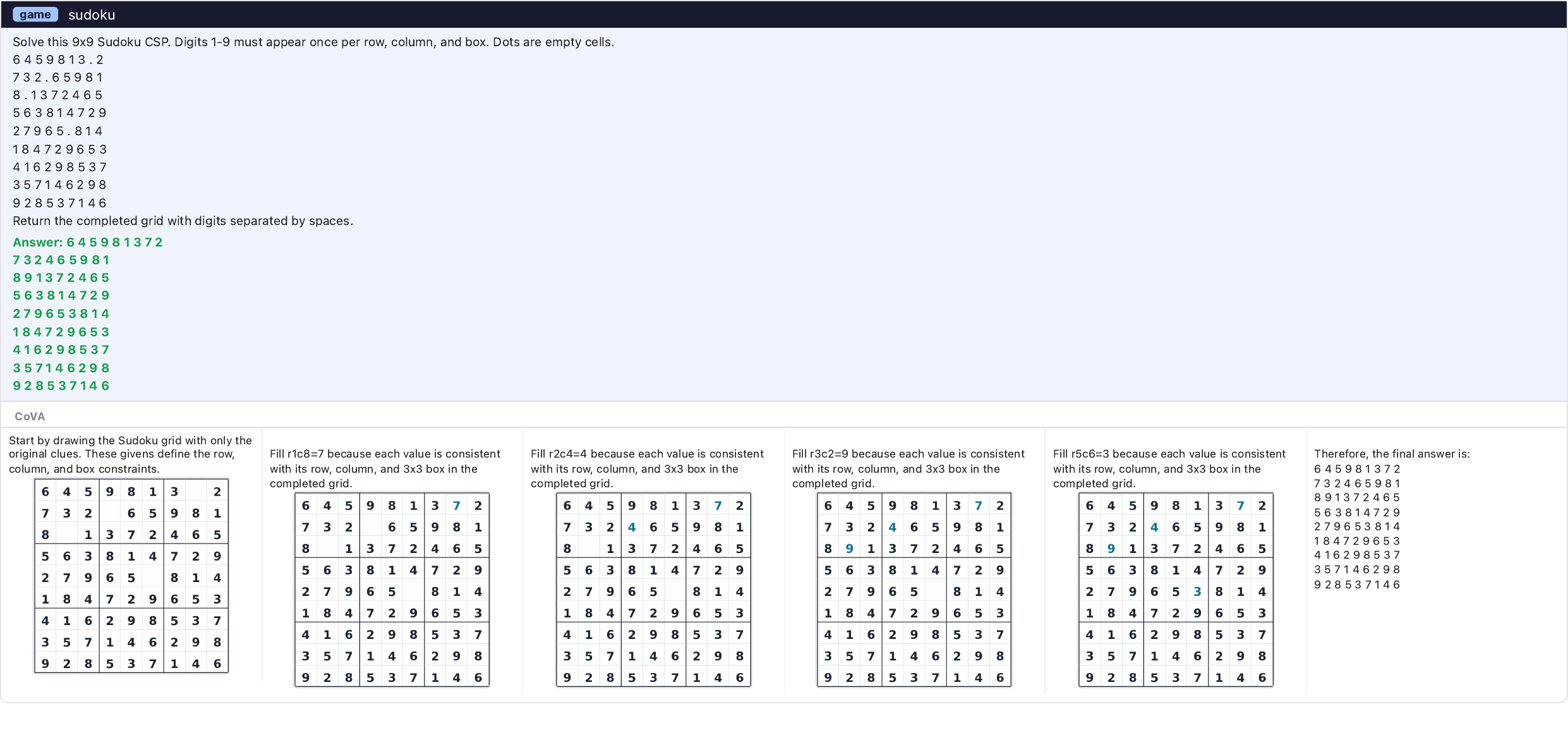}

    \includegraphics[
        width=\textwidth,
        clip,
        trim=0 6cm 0 0
    ]{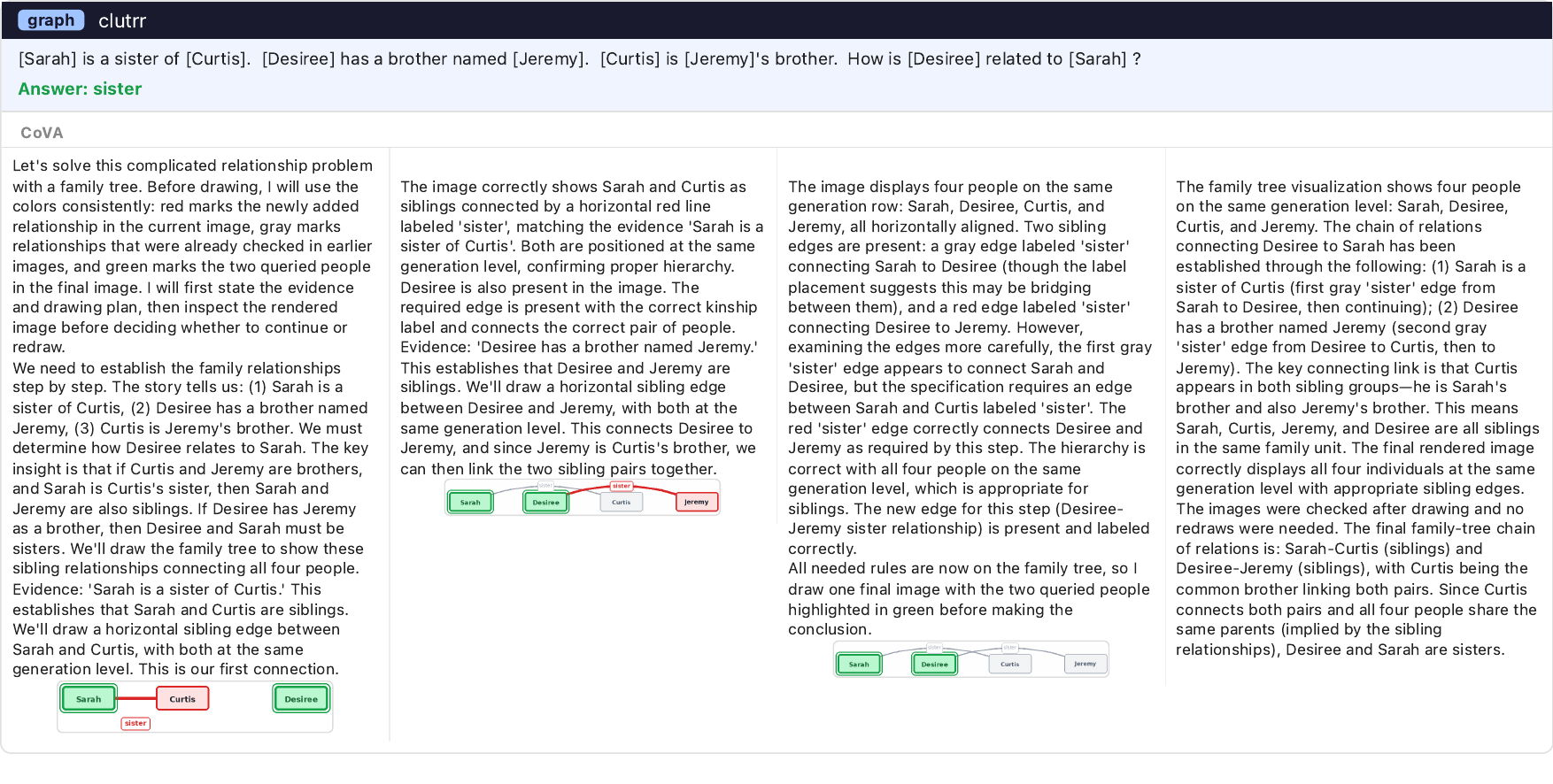}

    \includegraphics[
        width=\textwidth,
        clip,
        trim=0 13cm 0 0
    ]{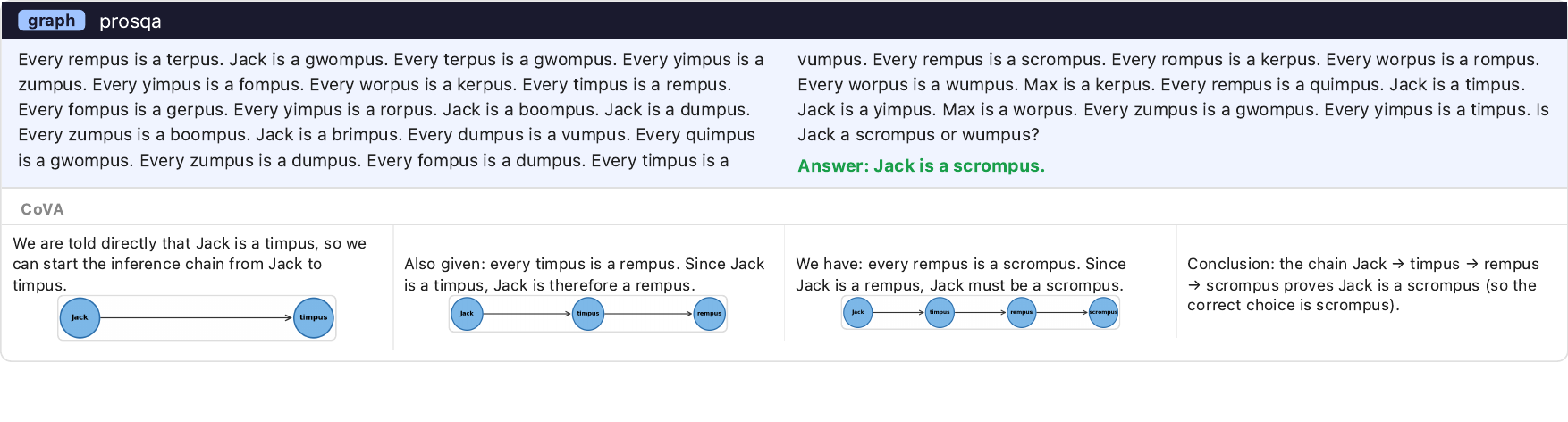}

    \caption[]{Training data examples (continued).}
\end{figure*}

\begin{figure*}[p]\ContinuedFloat
    \centering

    \includegraphics[
        width=\textwidth,
        clip,
        trim=0 8cm 0 0
    ]{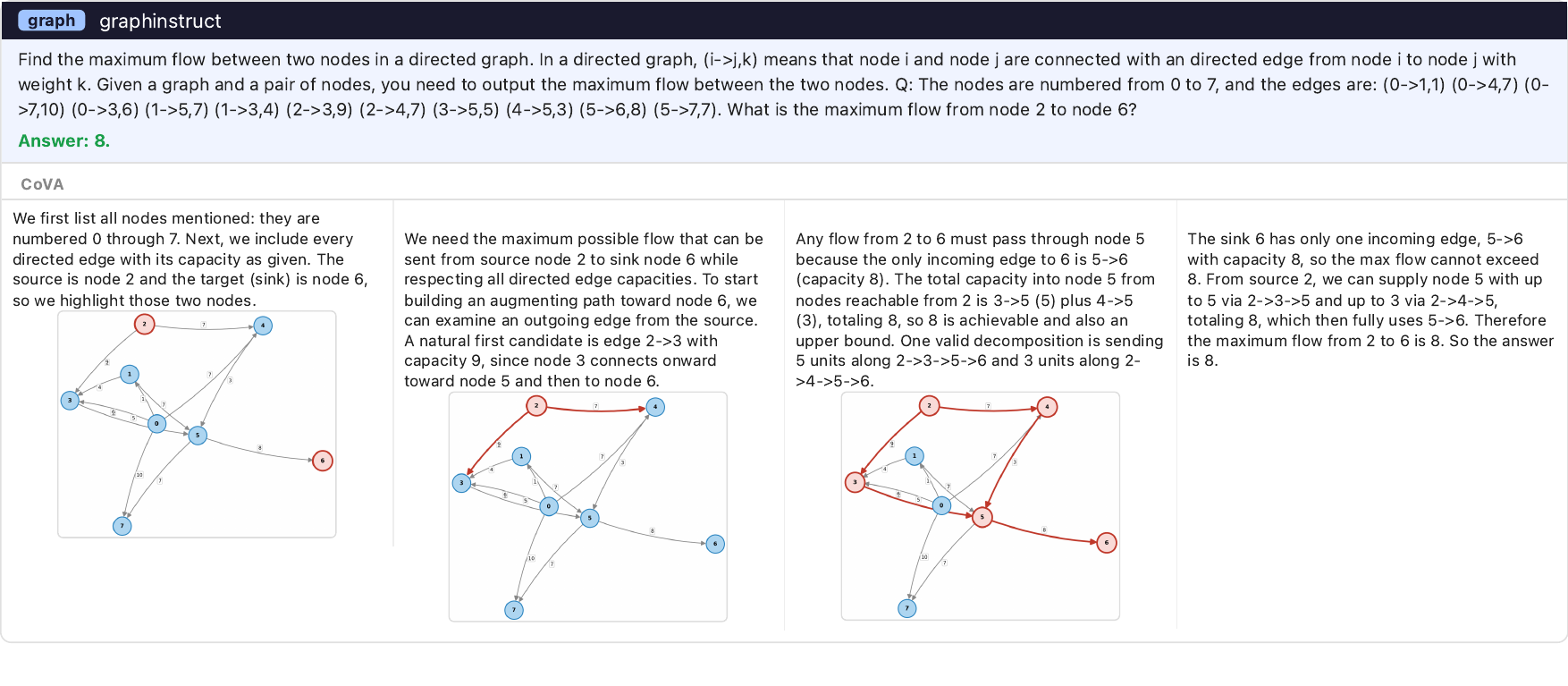}

    \includegraphics[
        width=\textwidth,
        clip,
        trim=0 6cm 0 0
    ]{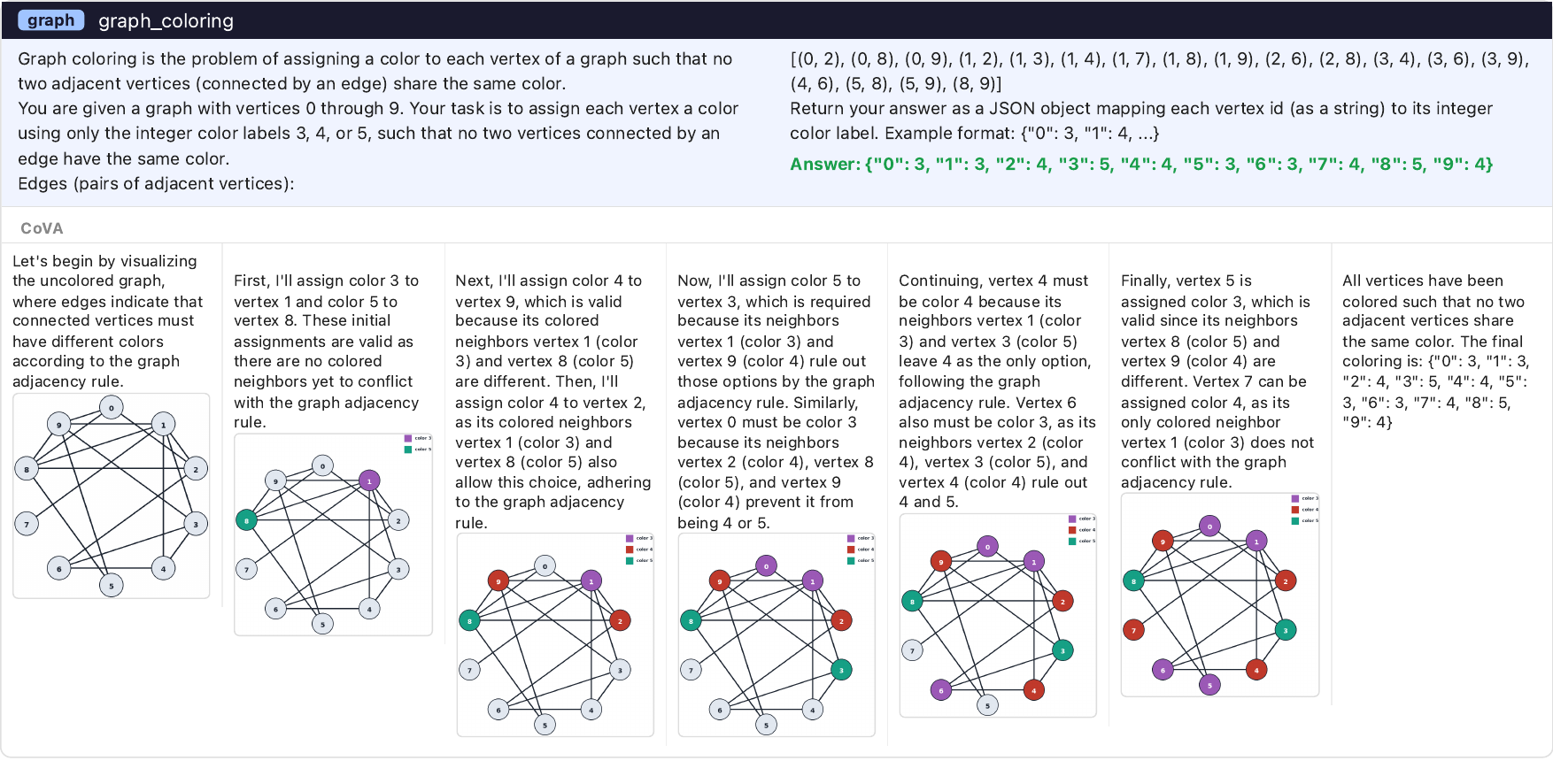}

    \caption[]{Training data examples (continued).}
\end{figure*}

\begin{figure*}[p]\ContinuedFloat
    \centering

    \includegraphics[
        width=\textwidth,
        clip,
        trim=0 12cm 0 0
    ]{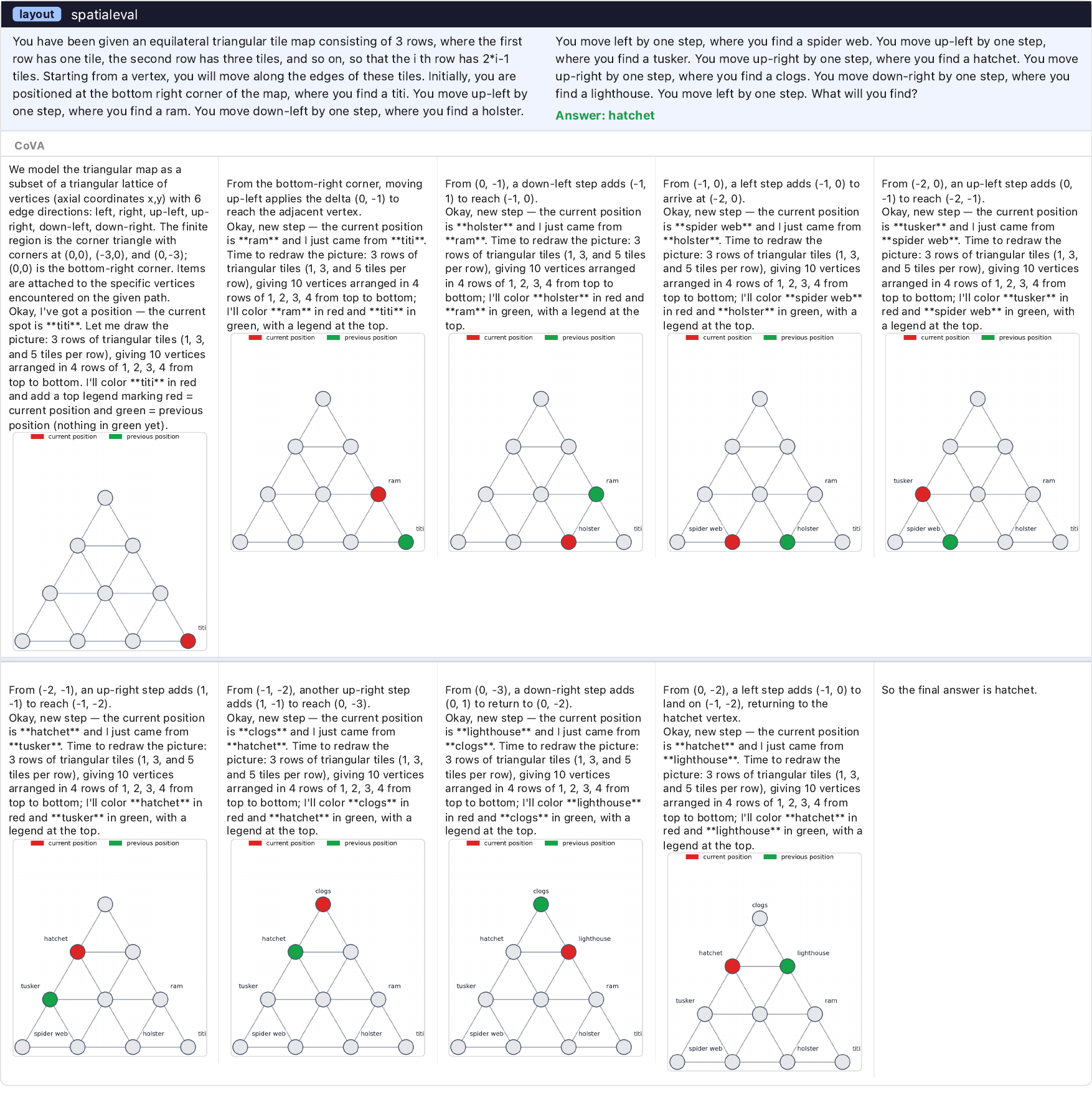}

    \includegraphics[
        width=\textwidth,
        clip,
        trim=0 10.5cm 0 0
    ]{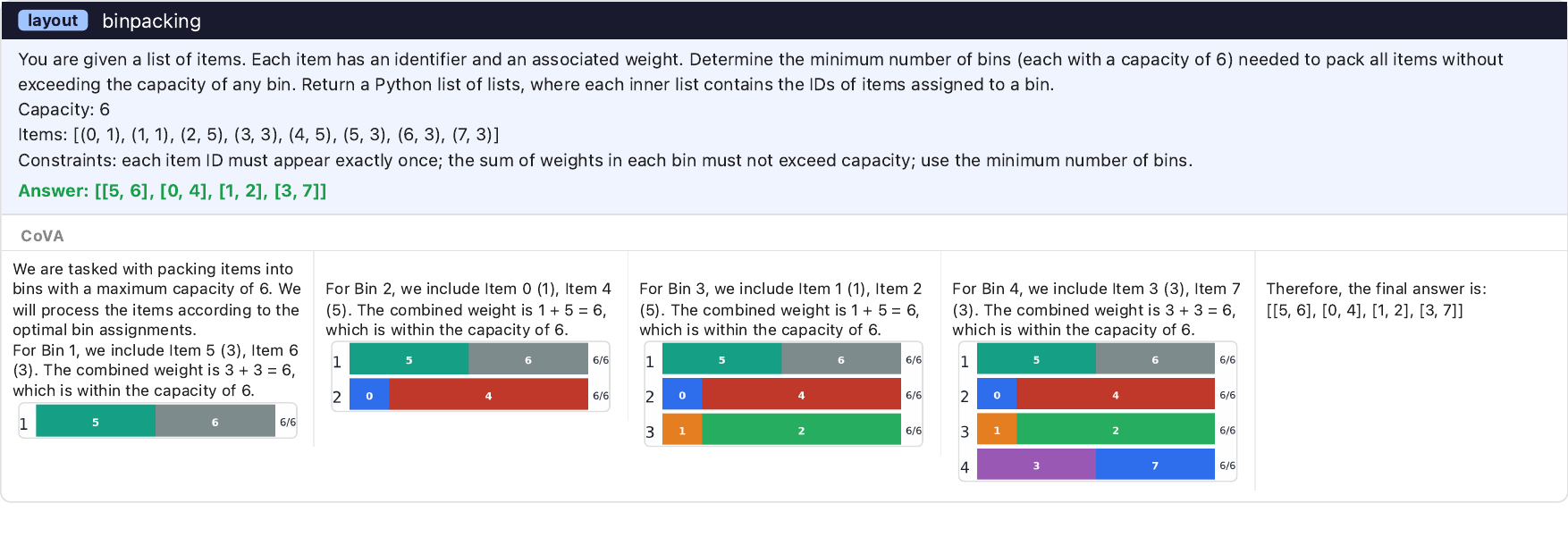}
    
    \caption[]{Training data examples (continued).}
\end{figure*}

\begin{figure*}[p]\ContinuedFloat
    \centering

    \includegraphics[
        width=\textwidth,
        clip,
        trim=0 11.5cm 0 0
    ]{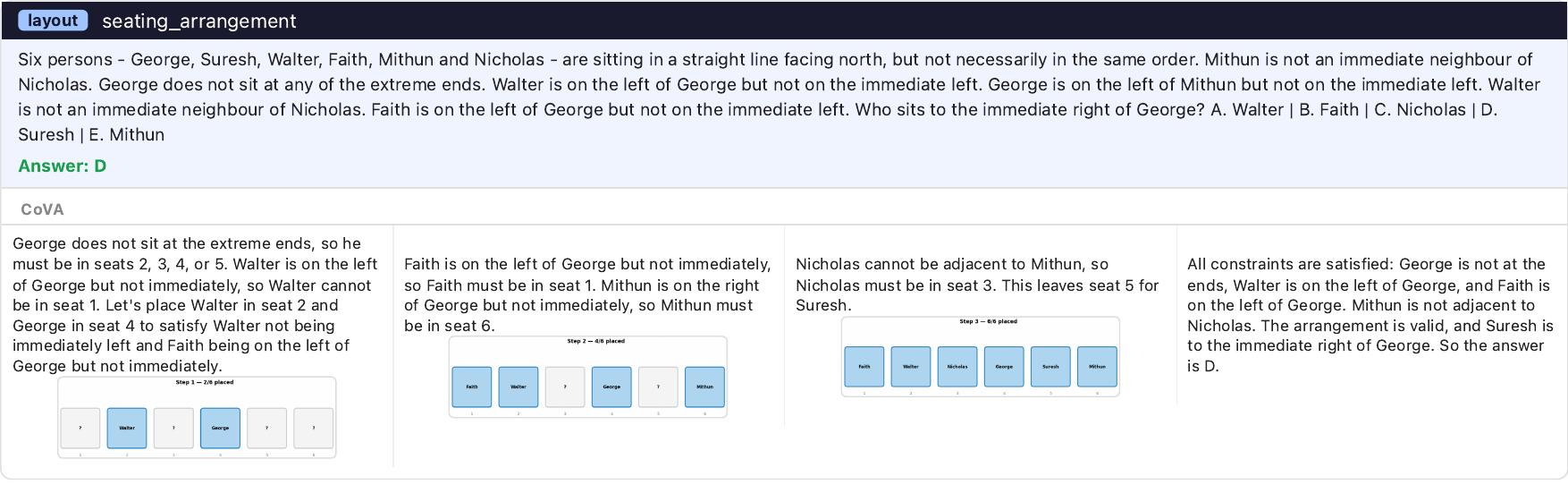}

    \includegraphics[
        width=\textwidth,
        clip,
        trim=0 3cm 0 0
    ]{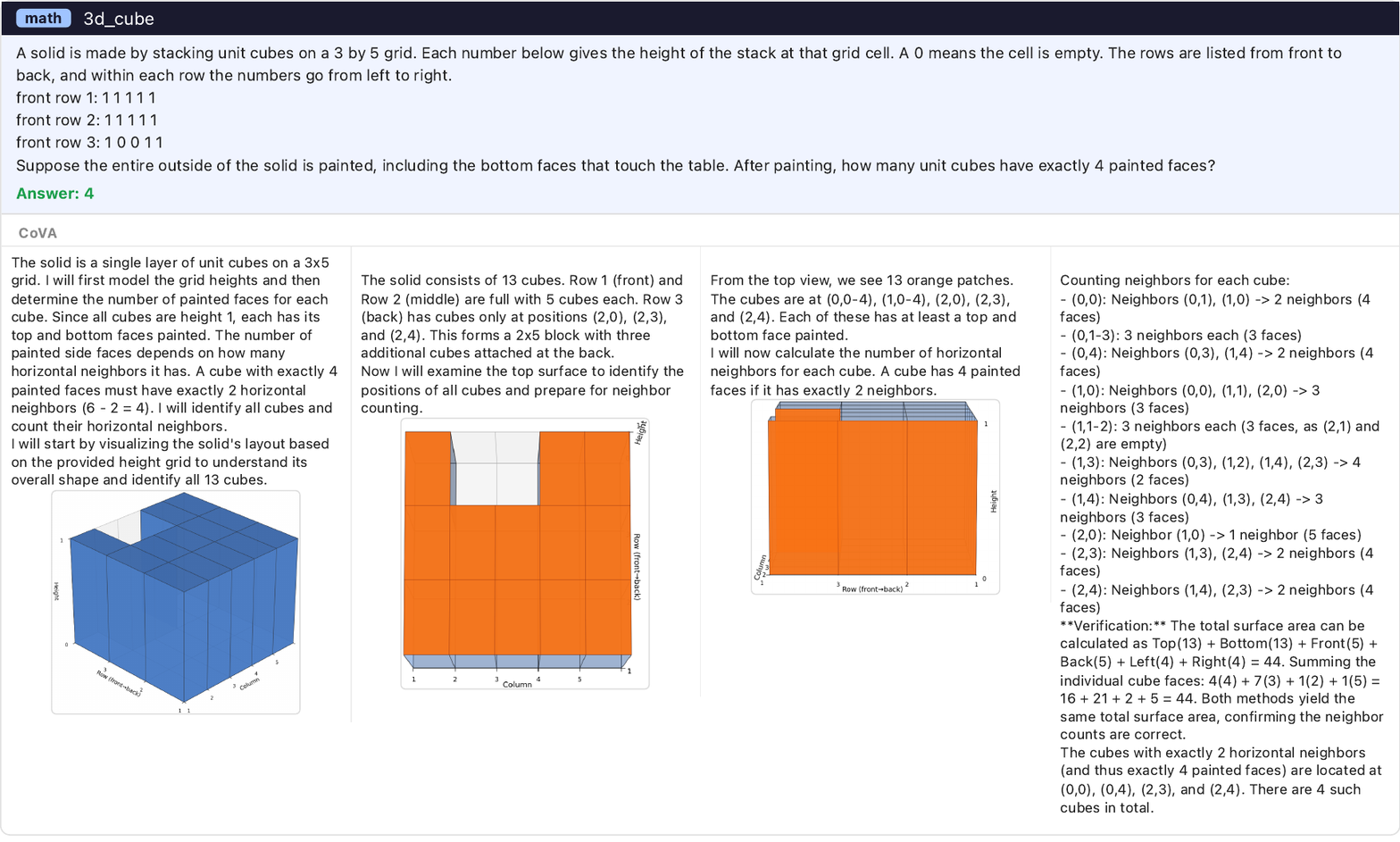}

    \includegraphics[
        width=\textwidth,
        clip,
        trim=0 7cm 0 0
    ]{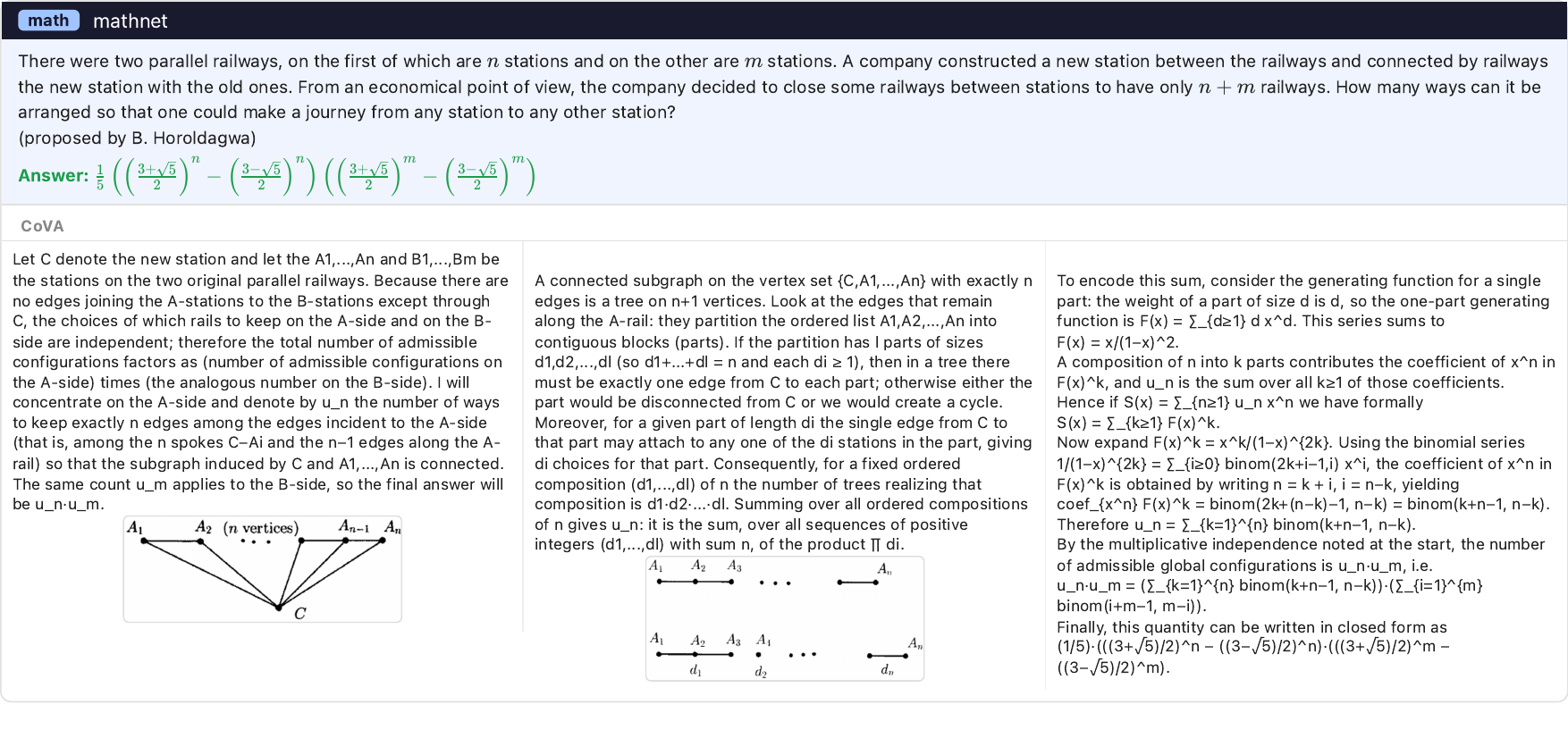}
    
    \caption[]{Training data examples (continued).}
\end{figure*}

\begin{figure*}[p]\ContinuedFloat
    \centering

    \includegraphics[
        width=\textwidth,
        clip,
        trim=0 6cm 0 0
    ]{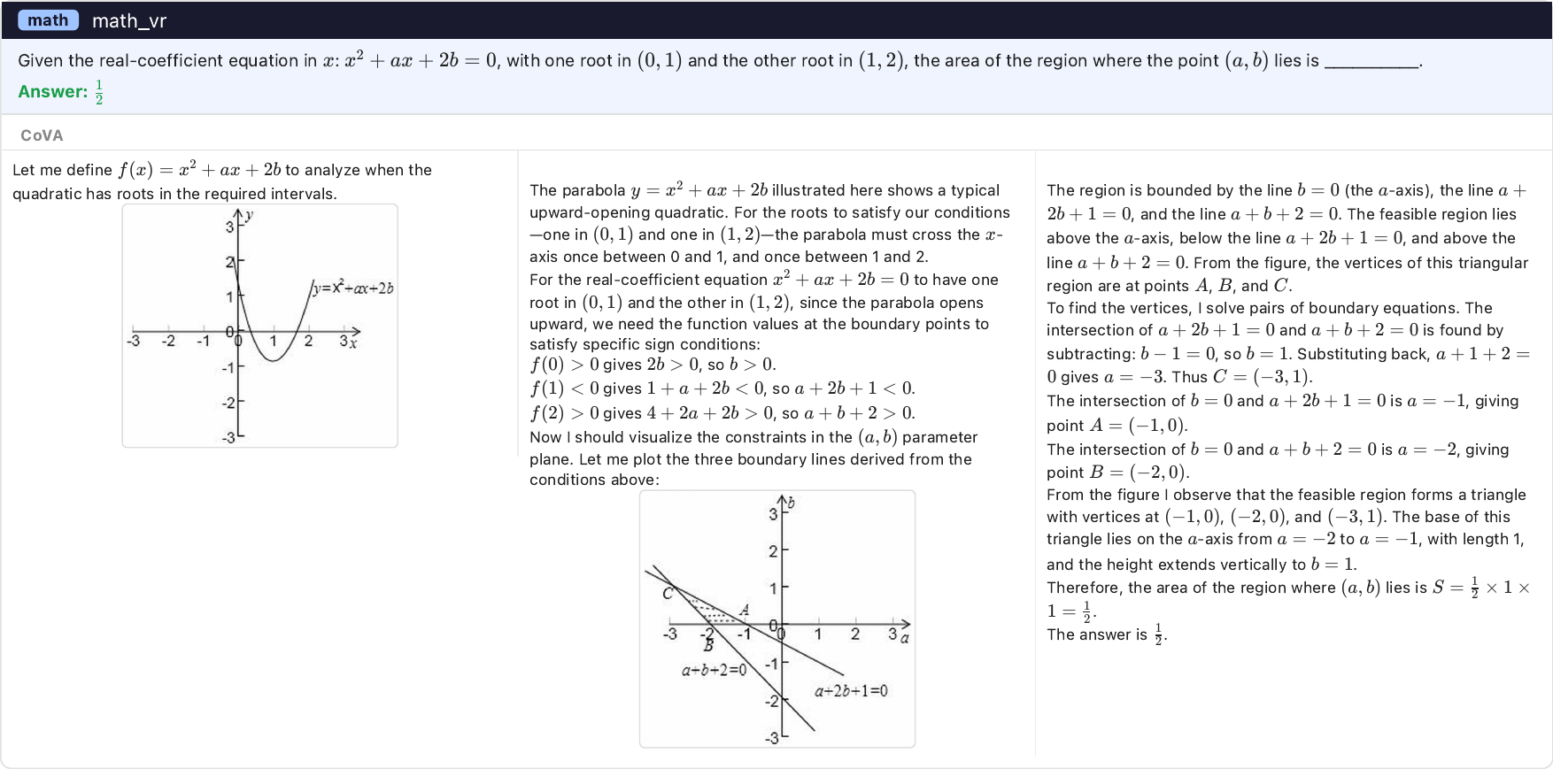}

    \includegraphics[
        width=\textwidth,
        clip,
        trim=0 7cm 0 0
    ]{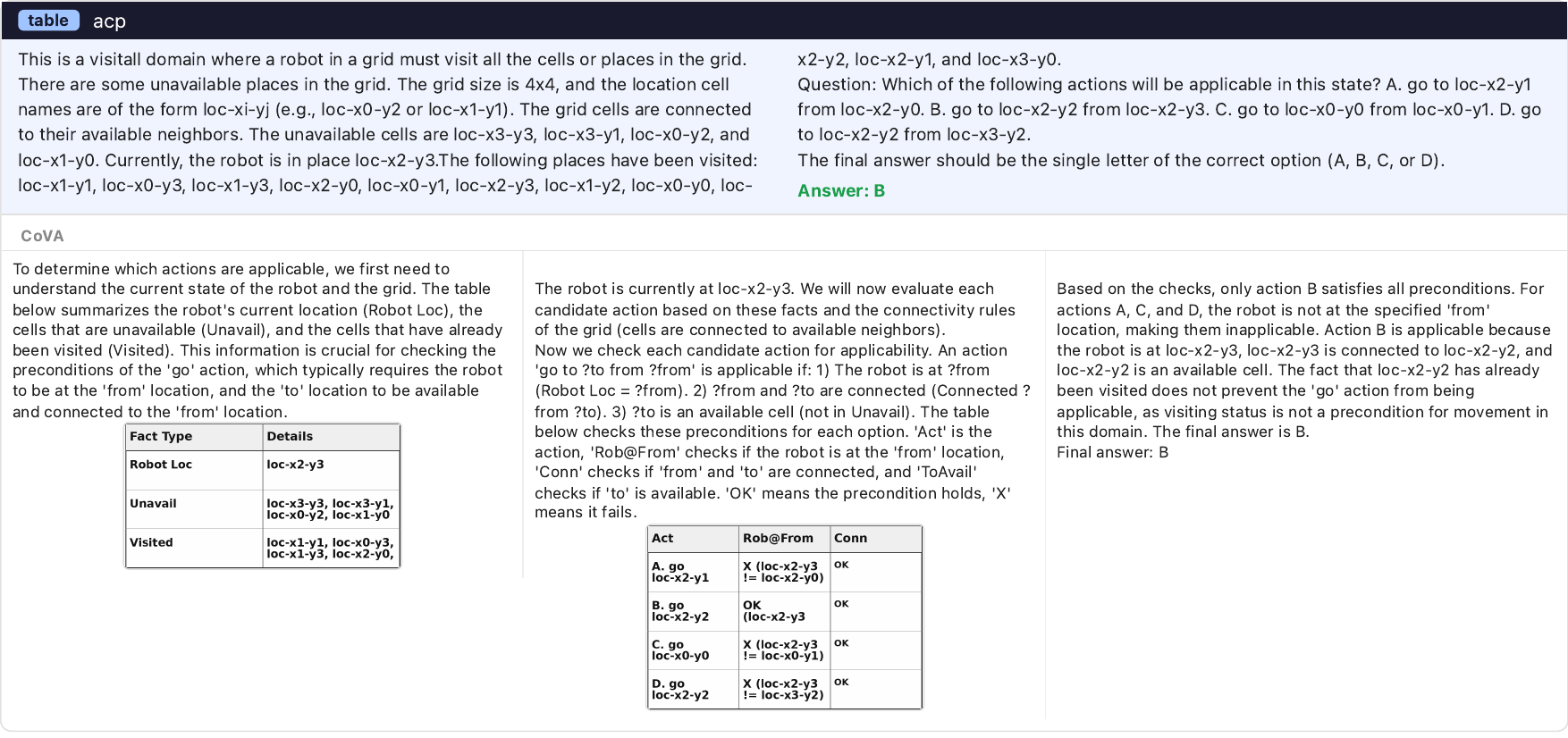}

    \includegraphics[
        width=\textwidth,
        clip,
        trim=0 5.5cm 0 0
    ]{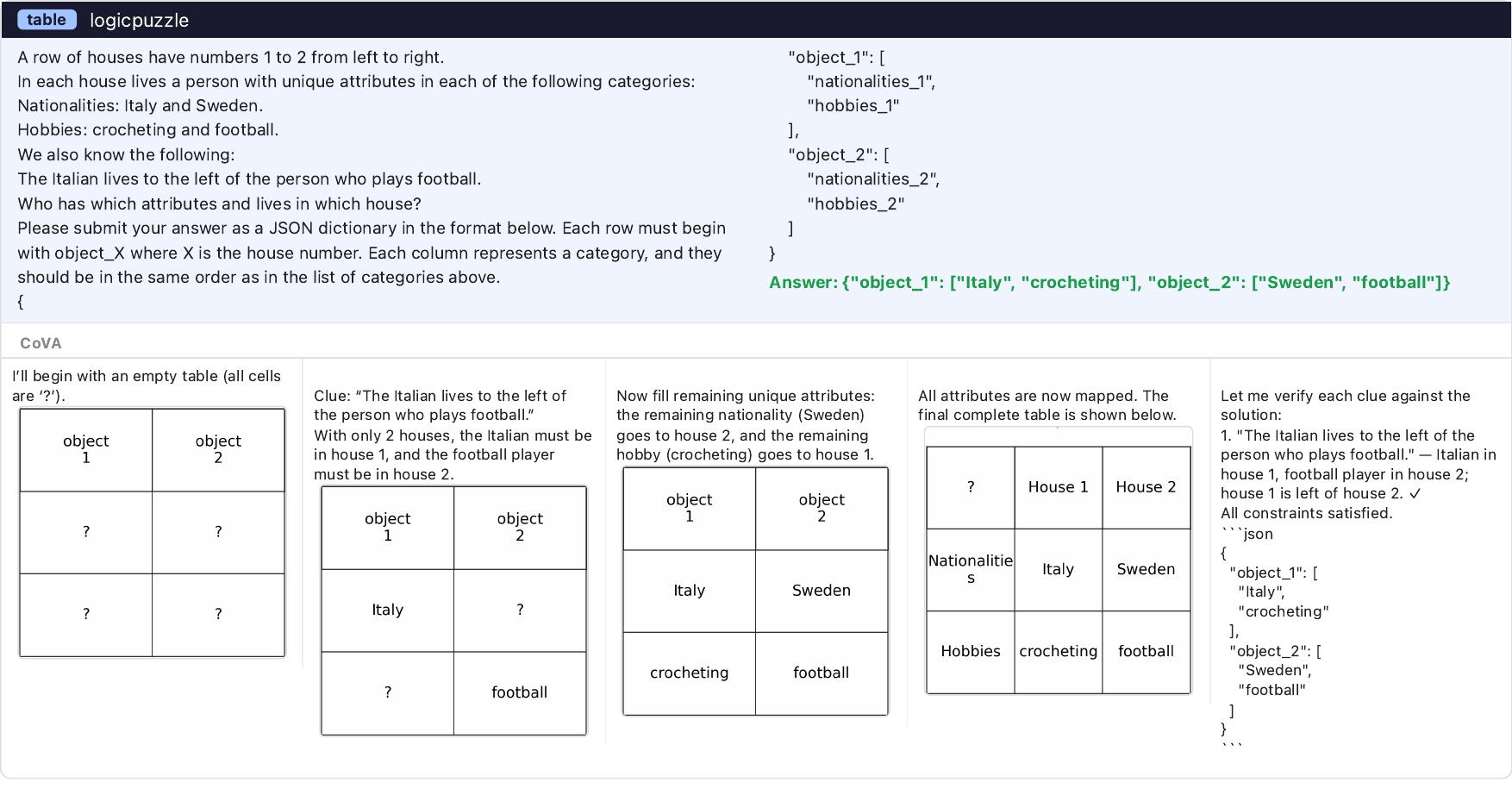}
    
    \caption[]{Training data examples (continued).}
\end{figure*}

\section{Ethics, Risks, and Artifact Documentation}
\label{app:ethics}

\paragraph{Potential risks.}
CoVA-SFT is intended as a research dataset for studying interleaved textual and visual reasoning. The dataset may inherit errors from the upstream vision-language model used to generate intermediate reasoning traces and rendered workspaces. Although our verification and self-correction loop is designed to reduce structural inconsistencies, subtle hallucinations, incorrect reasoning steps, or malformed renderings may remain in the released data. Models trained on such data may reproduce these errors or become overconfident in visually grounded reasoning traces. We therefore recommend that CoVA-SFT be used for research and evaluation rather than for high-stakes decision making without additional validation.

\paragraph{Artifact licenses and intended use.}
CoVA-SFT combines programmatically generated examples with examples derived from existing open-source datasets. The original portions of CoVA-SFT created by us will be released under the CC BY 4.0 License. For each external source used in the dataset, we list the original dataset name and URL in \autoref{tab:dataset_sources}; any examples derived from these sources remain subject to the applicable licenses and usage restrictions specified by their original creators. We use these artifacts for research purposes consistent with their intended use, including constructing and evaluating reasoning tasks involving graphs, games, layouts, tables, and mathematical reasoning. We do not intentionally include private, personally identifiable, or offensive content.

\section{Experimental Hyperparameters}
\label{sec:appendix_hyperparams}

This appendix provides additional details regarding the hyperparameter configurations and hardware infrastructure used during the training of CoVA.

\paragraph{Hardware and Infrastructure.} All experiments were conducted on a cluster of NVIDIA H200 (80GB) GPUs. SFT training was distributed across 8 H200 GPUs using Fully Sharded Data Parallel (FSDP2) in mixed-precision \texttt{bfloat16}. Training utilized FlashAttention v2 and Liger Kernel fused operations to optimize throughput and memory efficiency.

\paragraph{Optimization Details.} The hyperparameter settings are summarized in \autoref{tab:hyperparams}. During SFT we optimized the joint text and latent visual token objective. The visual encoder was unfrozen to allow for better representation alignment with the soft visual tokens. The training corpus of 51.9K interleaved samples was trained for 2 epochs.
\begin{table}[ht]
    \centering
    \small
    \renewcommand{\arraystretch}{1.2}
    \caption{Training hyperparameters for the SFT phase.}
    \label{tab:hyperparams}
    \begin{tabular}{lc}
        \toprule
        \textbf{Hyperparameter} & \textbf{SFT} \\
        \midrule
        Backbone Model & Qwen3-VL-8B-Thinking \\
        Optimizer & AdamW \\
        Learning Rate & $5 \times 10^{-5}$ \\
        LR Schedule & Cosine \\
        Warmup Steps & 50 \\
        Global Batch Size & 16 \\
        Epochs & 2 \\
        Visual Tokens per Image & 128 \\
        SFT Loss Weight ($\lambda$) & 1.0 \\
        Max Response Length & 32,768 \\
        \bottomrule
    \end{tabular}
\end{table}

\section{AI Use Disclosure}
\label{app:ai}

The authors used AI-based tools to assist with code generation, editing, and writing during the preparation of this paper. Specifically, AI assistance was used to help draft and revise portions of the manuscript for clarity, grammar, and organization, and to support the development, debugging, and refinement of code used in the research workflow. All AI-generated or AI-assisted content, code, analyses, and interpretations were reviewed, verified, and, where necessary, modified by the authors. The authors take full responsibility for the accuracy, integrity, originality, and final content of the paper, including any code or text developed with AI assistance.

%% file: custom.bib
@article{zhang2025openmmreasoner,
  title={Openmmreasoner: Pushing the frontiers for multimodal reasoning with an open and general recipe},
  author={Zhang, Kaichen and Wu, Keming and Yang, Zuhao and Li, Bo and Hu, Kairui and Wang, Bin and Liu, Ziwei and Li, Xingxuan and Bing, Lidong},
  journal={arXiv preprint arXiv:2511.16334},
  year={2025}
}

@article{jaech2024openai,
  title={Openai o1 system card},
  author={Jaech, Aaron and Kalai, Adam and Lerer, Adam and Richardson, Adam and El-Kishky, Ahmed and Low, Aiden and Helyar, Alec and Madry, Aleksander and Beutel, Alex and Carney, Alex and others},
  journal={arXiv preprint arXiv:2412.16720},
  year={2024}
}

@article{guo2025deepseek,
  title={Deepseek-r1: Incentivizing reasoning capability in llms via reinforcement learning},
  author={Guo, Daya and Yang, Dejian and Zhang, Haowei and Song, Junxiao and Wang, Peiyi and Zhu, Qihao and Xu, Runxin and Zhang, Ruoyu and Ma, Shirong and Bi, Xiao and others},
  journal={arXiv preprint arXiv:2501.12948},
  year={2025}
}

@article{deng2025openvlthinker,
  title={Openvlthinker: Complex vision-language reasoning via iterative sft-rl cycles},
  author={Deng, Yihe and Bansal, Hritik and Yin, Fan and Peng, Nanyun and Wang, Wei and Chang, Kai-Wei},
  journal={arXiv preprint arXiv:2503.17352},
  year={2025}
}

@article{qin2025chain,
  title={Chain-of-visual-thought: Teaching vlms to see and think better with continuous visual tokens},
  author={Qin, Yiming and Wei, Bomin and Ge, Jiaxin and Kallidromitis, Konstantinos and Fu, Stephanie and Darrell, Trevor and Wang, XuDong},
  journal={arXiv preprint arXiv:2511.19418},
  year={2025}
}

@article{hu2024visual,
  title={Visual sketchpad: Sketching as a visual chain of thought for multimodal language models},
  author={Hu, Yushi and Shi, Weijia and Fu, Xingyu and Roth, Dan and Ostendorf, Mari and Zettlemoyer, Luke and Smith, Noah A and Krishna, Ranjay},
  journal={Advances in Neural Information Processing Systems},
  volume={37},
  pages={139348--139379},
  year={2024}
}

@inproceedings{menon2024whiteboard,
  title={Whiteboard-of-thought: Thinking step-by-step across modalities},
  author={Menon, Sachit and Zemel, Richard and Vondrick, Carl},
  booktitle={Proceedings of the 2024 Conference on Empirical Methods in Natural Language Processing},
  pages={20016--20031},
  year={2024}
}

@article{su2025openthinkimg,
  title={Openthinkimg: Learning to think with images via visual tool reinforcement learning},
  author={Su, Zhaochen and Li, Linjie and Song, Mingyang and Hao, Yunzhuo and Yang, Zhengyuan and Zhang, Jun and Chen, Guanjie and Gu, Jiawei and Li, Juntao and Qu, Xiaoye and others},
  journal={arXiv preprint arXiv:2505.08617},
  year={2025}
}

@article{li2025imagine,
  title={Imagine while reasoning in space: Multimodal visualization-of-thought},
  author={Li, Chengzu and Wu, Wenshan and Zhang, Huanyu and Xia, Yan and Mao, Shaoguang and Dong, Li and Vuli{\'c}, Ivan and Wei, Furu},
  journal={arXiv preprint arXiv:2501.07542},
  year={2025}
}

@inproceedings{bigverdi2025perception,
  title={Perception tokens enhance visual reasoning in multimodal language models},
  author={Bigverdi, Mahtab and Luo, Zelun and Hsieh, Cheng-Yu and Shen, Ethan and Chen, Dongping and Shapiro, Linda G and Krishna, Ranjay},
  booktitle={Proceedings of the Computer Vision and Pattern Recognition Conference},
  pages={3836--3845},
  year={2025}
}

@article{yang2025machine,
  title={Machine mental imagery: Empower multimodal reasoning with latent visual tokens},
  author={Yang, Zeyuan and Yu, Xueyang and Chen, Delin and Shen, Maohao and Gan, Chuang},
  journal={arXiv preprint arXiv:2506.17218},
  year={2025}
}

@article{li2025latent,
  title={Latent visual reasoning},
  author={Li, Bangzheng and Sun, Ximeng and Liu, Jiang and Wang, Ze and Wu, Jialian and Yu, Xiaodong and Chen, Hao and Barsoum, Emad and Chen, Muhao and Liu, Zicheng},
  journal={arXiv preprint arXiv:2509.24251},
  year={2025}
}

@article{wang2025autoregressive,
  title={Autoregressive semantic visual reconstruction helps vlms understand better},
  author={Wang, Dianyi and Song, Wei and Wang, Yikun and Wang, Siyuan and Yu, Kaicheng and Wei, Zhongyu and Wang, Jiaqi},
  journal={arXiv preprint arXiv:2506.09040},
  year={2025}
}

@article{li2025zebracot,
  title={Zebra-CoT: A Dataset for Interleaved Vision Language Reasoning},
  author={Li, Ang and Wang, Charles and Yue, Kaiyu and Cai, Zikui and Liu, Ollie and Fu, Deqing and Guo, Peng and Zhu, Wang Bill and Sharan, Vatsal and Jia, Robin and Neiswanger, Willie and Huang, Furong and Goldstein, Tom and Goldblum, Micah},
  journal={arXiv preprint arXiv:2507.16746},
  year={2025}
}

@article{chern2025thinking,
  title={Thinking with generated images},
  author={Chern, Ethan and Hu, Zhulin and Chern, Steffi and Kou, Siqi and Su, Jiadi and Ma, Yan and Deng, Zhijie and Liu, Pengfei},
  journal={arXiv preprint arXiv:2505.22525},
  year={2025}
}

@article{shi2025mathcanvas,
  title={Mathcanvas: Intrinsic visual chain-of-thought for multimodal mathematical reasoning},
  author={Shi, Weikang and Yu, Aldrich and Fang, Rongyao and Ren, Houxing and Wang, Ke and Zhou, Aojun and Tian, Changyao and Fu, Xinyu and Hu, Yuxuan and Lu, Zimu and others},
  journal={arXiv preprint arXiv:2510.14958},
  year={2025}
}

@article{duan2025codeplot,
  title={Codeplot-cot: Mathematical visual reasoning by thinking with code-driven images},
  author={Duan, Chengqi and Sun, Kaiyue and Fang, Rongyao and Zhang, Manyuan and Feng, Yan and Luo, Ying and Liu, Yufang and Wang, Ke and Pei, Peng and Cai, Xunliang and others},
  journal={arXiv preprint arXiv:2510.11718},
  year={2025}
}

@article{yang2025qwen3,
  title={Qwen3 technical report},
  author={Yang, An and Li, Anfeng and Yang, Baosong and Zhang, Beichen and Hui, Binyuan and Zheng, Bo and Yu, Bowen and Gao, Chang and Huang, Chengen and Lv, Chenxu and others},
  journal={arXiv preprint arXiv:2505.09388},
  year={2025}
}

@article{bai2025qwen3,
  title={Qwen3-vl technical report},
  author={Bai, Shuai and Cai, Yuxuan and Chen, Ruizhe and Chen, Keqin and Chen, Xionghui and Cheng, Zesen and Deng, Lianghao and Ding, Wei and Gao, Chang and Ge, Chunjiang and others},
  journal={arXiv preprint arXiv:2511.21631},
  year={2025}
}
